\pdfoutput=1
\documentclass[11pt]{article}

\usepackage[final]{acl}
\usepackage{times}
\usepackage{latexsym}
\usepackage{graphicx}
\usepackage[T1]{fontenc}
\usepackage[utf8]{inputenc}
\usepackage{amsmath}
\usepackage{changepage}
\usepackage{stix,bbding,pifont,utfsym,fontawesome}
\usepackage{adjustbox}
\usepackage{rotating}
\usepackage{array}
\usepackage{arydshln}
\usepackage{booktabs}
\usepackage{multirow}
\usepackage{tabularx}
\usepackage{colortbl}

\newcolumntype{D}{>{\dottedcolumn}c<{\enddottedcolumn}}

\title{Adapting Multilingual LLMs to Low-Resource Languages with Knowledge Graphs via Adapters}

\author{Daniil Gurgurov\textsuperscript{\normalfont1,2} \quad Mareike Hartmann\textsuperscript{\normalfont2} \quad Simon Ostermann\textsuperscript{\normalfont1} \\
  \textsuperscript{1}German Research Center for Artificial Intelligence (DFKI) \\ \textsuperscript{2}Department of Language Science and Technology, Saarland University \\
  {\small \texttt{daniil.gurgurov@dfki.de, mareikeh@coli.uni-saarland.de, simon.ostermann@dfki.de} }
}

\begin{document}
\maketitle

\begin{abstract}
This paper explores the integration of graph knowledge from linguistic ontologies into multilingual Large Language Models (LLMs) using adapters to improve performance for low-resource languages (LRLs) in sentiment analysis (SA) and named entity recognition (NER). Building upon successful parameter-efficient fine-tuning techniques, such as K-ADAPTER \cite{wang2020k} and MAD-X \cite{pfeiffer2020mad}, we propose a similar approach for incorporating knowledge from multilingual graphs, connecting concepts in various languages with each other through linguistic relationships, into multilingual LLMs for LRLs. Specifically, we focus on eight LRLs —Maltese, Bulgarian, Indonesian, Nepali, Javanese, Uyghur, Tibetan, and Sinhala — and employ language-specific adapters fine-tuned on data extracted from the language-specific section of ConceptNet, aiming to enable knowledge transfer across the languages covered by the knowledge graph. We compare various fine-tuning objectives, including standard Masked Language Modeling (MLM), MLM with full-word masking, and MLM with targeted masking, to analyze their effectiveness in learning and integrating the extracted graph data. Through empirical evaluation on language-specific tasks, we assess how structured graph knowledge affects the performance of multilingual LLMs for LRLs in SA and NER, providing insights into the potential benefits of adapting language models for low-resource scenarios.
\end{abstract}

\section{Introduction}
In recent years, the advancement of multilingual Large Language Models (LLMs) \cite{devlin2018bert, conneau2019unsupervised, xue2020mt5} has revolutionized the field of natural language processing (NLP), enabling impressive performance across various languages. However, these models often struggle with low-resource languages (LRLs), where limited data availability affects their effectiveness \cite{wu-dredze-2020-languages}. To address this limitation, researchers have explored integrating external knowledge sources into multilingual LLMs to enhance their performance in both high-resource and low-resource contexts \cite{wang2020k, lauscher2020common, pfeiffer2020mad} via adapters \cite{houlsby2019parameter} and full fine-tuning. 

Adapters, introduced by \citet{houlsby2019parameter}, are small modules inserted between the layers of a model and trained while the model is kept frozen. Previous work has used such Adapters to integrate external knowledge into LLMs. For instance, \citet{wang2020k} demonstrated improvements in relation classification, entity typing, and question answering tasks by integrating graph knowledge from Wikidata \cite{vrandevcic2014wikidata} into RoBERTa \cite{liu2019roberta} using adapters. Similarly, \citet{lauscher2020common} enhanced BERT \cite{devlin2018bert} with graph knowledge from ConceptNet \cite{speer2017conceptnet}, achieving significant performance gains on tasks requiring common-sense knowledge. However, these efforts primarily focused on the English language. In contrast, \citet{pfeiffer2020mad} addressed low-resource languages by integrating textual knowledge from Wikipedia into XLM-R \cite{conneau2019unsupervised} via language adapters. Their approach demonstrated improvements over the baseline model for named entity recognition (NER) task. 

Motivated by recent advancements in the integration of graph knowledge into language models, particularly for English, this paper investigates the incorporation of graph knowledge from linguistic ontologies, specifically ConceptNet, into multilingual LLMs particularly for LRLs. Injecting such data might be beneficial due to the scarcity of training data for these languages and the additional semantic and multilingual information provided by knowledge graphs \cite{miller1995wordnet, speer2017conceptnet}. Our focus is on a subset of LRLs, aiming to extend the success observed in graph knowledge integration to linguistically diverse and resource-scarce contexts. We work with Maltese, Bulgarian, Indonesian, Nepali, Javanese, Uyghur, Tibetan, and Sinhala, identified as low-resource according to \citeauthor{joshi2020state} (\citeyear{joshi2020state}). Our primary objective is to evaluate whether injecting multilingual graph knowledge, connecting various languages through linguistic relationships, into pre-trained multilingual LLMs through adapters improves performance for LRLs. We train language-specific adapters on ConceptNet data using different objective functions, including standard Masked Language Modeling (MLM) \cite{devlin2018bert}, MLM with full-word masking \cite{cui2021pre}, and MLM with targeted masking, and evaluate the downstream performance of the adapted model on sentiment analysis (SA) and NER tasks. 

Our work extends existing advancements by proposing an approach that utilizes adapters to integrate graph knowledge specifically for LRLs, following a modular design similar to the one introduced by \citeauthor{pfeiffer2020mad} (\citeyear{pfeiffer2020mad}). Our contributions include:
\begin{itemize}
    \item \textbf{Low-Resource Languages Focus:} Unlike prior works on graph knowledge integration \cite{lauscher2020common, wang2020k}, our research concentrates explicitly on improving multilingual LLMs through the external graph knowledge injection for low-resource scenarios.
    \item \textbf{Exploiting Various Knowledge Sources and Types:} We investigate the integration of language adapters based on Wikipedia and ConceptNet, both individually and in combination. This expands the approach of \citeauthor{pfeiffer2020mad} (\citeyear{pfeiffer2020mad}), which solely utilized Wikipedia data, enabling a comprehensive assessment of different knowledge sources' impact on model performance.  We assume that language models can benefit from the multilingual connections in ConceptNet.
    \item \textbf{Single-Language Training Approach:} In contrast to the multilingual transfer learning approach used by \citeauthor{pfeiffer2020mad} (\citeyear{pfeiffer2020mad}), which primarily focuses on cross-lingual adaptation, our methodology involves training language and task adapters using data in the same LRL. This training strategy aims to maximize model performance and adaptability to the specific linguistic characteristics of each target language.
\end{itemize}

\begin{figure}[h]
    \centering
    \includegraphics[width=0.35\textwidth]{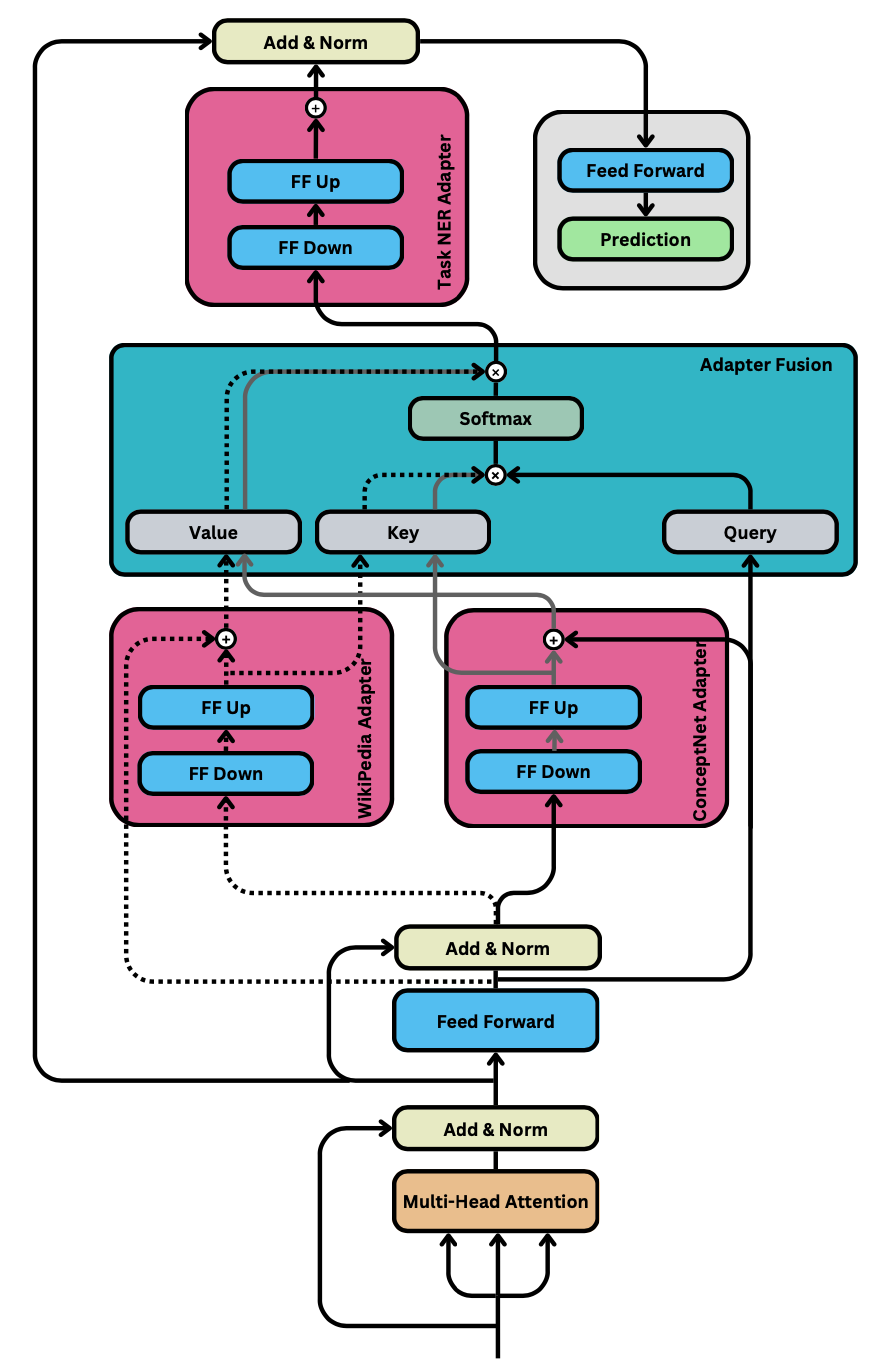}
    \caption{Proposed method. One of the Wiki or ConceptNet language adapters is used during inference. The outputs then go to a task adapter, which is followed by a classification head. If fusion is specified, the fusion mechanism is activated.}
    \label{fig:method}
\end{figure}

Our study provides insights into the potential benefits of adapting multilingual LLMs for low resource scenarios, contributing to the ongoing exploration of multilingual language model adaptation and graph knowledge integration.

The complete code for our experiments is publicly available on GitHub\footnote{\url{https://github.com/d-gurgurov/Injecting-Commonsense-Knowledge-into-LLMs}}.

\section{Related Work}
Our work extends recent advancements in integrating external graph knowledge into pre-trained LLMs \cite{lauscher2020common, wang2020k} and adapting pre-trained multilingual LLMs to specific languages \cite{pfeiffer2020mad}. We use the methodologies proposed in these studies as a foundation for enhancing the performance of multilingual LLMs on downstream tasks, particularly for LRLs. To this end, we provide the overview of these methods as well as other strategies of adapting multilingual LLMs to LRLs \cite{artetxe-etal-2020-cross, muller-etal-2021-unseen, vernikos-popescu-belis-2021-subword-mapping, pfeiffer-etal-2022-lifting}.

\subsection{Adapter-based Knowledge Integration}
In our approach, we draw inspiration from recently proposed adapter-based knowledge integration techniques, particularly the concept of K-Adapters \cite{wang2020k} and the work by \citet{lauscher2020common}. K-Adapters introduced a novel approach for injecting knowledge into pre-trained models like RoBERTa \cite{liu2019roberta} without modifying their original parameters. This method utilizes two types of adapters dedicated to factual and linguistic knowledge, demonstrating improvements in tasks such as entity typing and question answering. Factual adapters are trained with a relation classification objective using data aligned from Wikipedia text to Wikidata triplets \cite{vrandevcic2014wikidata}, while linguistic adapters are trained with a dependency relation prediction objective using linguistic information obtained from available dependency parsers.

Similarly, \citet{lauscher2020common} explored injecting external knowledge, specifically from ConceptNet \cite{speer2017conceptnet} and the Open Mind Common Sense (OMCS) corpus \cite{singh2002open}, into language models. They introduced two boosted models: CN-ADAPT and OM-ADAPT. CN-ADAPT involves creating a synthetic corpus through random traversal of the ConceptNet graph, with adapter parameters learned through Masked Language Modeling (MLM) training on this synthetic corpus. In OM-ADAPT, adapter parameters are learned directly through MLM training on the OMCS corpus. Both models employ a parameter-efficient adapter-based architecture \cite{houlsby2019parameter}, injecting bottleneck adapters into BERT's \cite{devlin2018bert} transformer layers.

An approach similar to ours is presented in \citet{hou2022adapters}, who however make stronger assumptions on the training data (such as the alignment of entities in the data with a knowledge graph) and rather train models to represent entities explicitly.

\begin{table*}[htb]
    \centering
    \small
    \begin{tabular}{l l}
        \toprule
        \textbf{ConceptNet Relationship} & \textbf{Natural Language Predicate} \\
        \midrule
        Antonym & is the opposite of \\
        DerivedFrom & is derived from \\
        EtymologicallyDerivedFrom & is etymologically derived from \\
        EtymologicallyRelatedTo & is etymologically related to \\
        FormOf & is a form of \\
        HasContext & has context of \\
        IsA & is a type of \\
        RelatedTo & is related to \\
        SimilarTo & is similar to \\
        Synonym & is a synonym of \\
        SymbolOf & is a symbol of \\
        DistinctFrom & is distinct from \\
        \bottomrule
    \end{tabular}
    \caption{Predefined mapping from ConceptNet relations to natural language predicates used for training ConceptNet-based Language Adapters.}
    \label{tab:relationship_mapping}
\end{table*}

\subsection{Language Adapters}
In our exploration of injecting graph knowledge for LRL scenarios, we follow the MAD-X architecture presented by \citeauthor{pfeiffer2020mad} (\citeyear{pfeiffer2020mad}). MAD-X offers an efficient approach to adapt pre-trained language models to LRLs by utilizing a modular structure consisting of language adapters, task-specific adapters, and invertible adapters.

The MAD-X framework utilizes language adapters as a fundamental component to adapt the model to specific languages. These adapters are trained on language-specific Wikipedia data and stacked onto the pre-trained model, allowing the model to capture language-specific nuances and patterns effectively. Following the enhanced bottleneck architecture \cite{pfeiffer2020adapterfusion}, the language adapter involves down- and up-projections with ReLU activation and is trained via MLM on unlabeled data.

During downstream task training, such as named entity recognition (NER), the fixed language adapter corresponding to the source language is used, ensuring adaptability to different languages without changing the underlying multilingual model. The embeddings are passed through the fixed language adapter before entering the task adapter, facilitating efficient adaptation to diverse linguistic contexts and specific task requirements.

Additionally, MAD-X introduces invertible adapters to mitigate the mismatch between multilingual and target language vocabulary. These invertible adapters are stacked on top of the embedding layer, with their respective inverses preceding the output embedding layer.

Task-specific adapters are then stacked on top of the language and invertible adapters to capture task-specific knowledge and specialize a language model in a certain task. 

These insights from the MAD-X framework serve as a valuable reference for our research on injecting structured graph knowledge into multilingual LLMs for low-resource cases.

\subsection{Adapting LLMs to Low-Resource Languages}
Various other strategies have been proposed to address the challenges of adapting multilingual LLMs to LRLs, particularly for languages with limited pre-training data. \citet{pfeiffer-etal-2022-lifting} propose X-MOD, a modular multilingual architecture that integrates shared and language-specific parameters to overcome the curse of multilinguality \cite{conneau2019unsupervised}, allowing efficient handling of linguistic diversity and supporting the extension to new languages with minimal performance impact on pre-trained languages. Additionally, \citet{artetxe-etal-2020-cross} train a new embedding layer with a corresponding target-language tokenizer to extend monolingual models to new languages, aiding language extension while maintaining model stability. Moreover, approaches based on transliteration and subword mappings have been proposed to incorporate additional languages into multilingual models, contributing to the expansion of multilingual capabilities of LLMs \cite{muller-etal-2021-unseen, vernikos-popescu-belis-2021-subword-mapping}. \citet{hangya-etal-2022-improving} present a bootstrapping-based approach for enhancing low-resource languages in multilingual LLMs, which relies on unsupervised word translation pairs from monolingual corpora.

\section{Injecting External Knowledge into LLMs for LRLs}
This section describes our approach for enhancing multilingual LLMs for LRLs by injecting external knowledge. We discuss the use of language adapters trained on ConceptNet and Wikipedia data, explore Adapter Fusion \cite{pfeiffer2020adapterfusion} for combining knowledge sources, and describe task adapters for fine-tuning multilingual LLMs for specific tasks. Our proposed method is illustrated in Figure \ref{fig:method}.

\subsection{Language Adapters}
We use language adapters for integrating external knowledge into multilingual LLMs and adapting these models to a specific language. In our study, two types of language adapters are employed: those trained on ConceptNet data and those trained on Wikipedia.


\subsubsection{ConceptNet Data Preparation}
ConceptNet-based language adapters are trained on knowledge extracted from ConceptNet, providing a rich source of linguistic relationships and semantic information across various languages.
Data preparation involves retrieving and formatting data from ConceptNet and converting it into natural text\footnote{For the extraction process, we utilized a dedicated self-built module for fetching data from CN, built on top of the CN API\ (\url{https://github.com/commonsense/conceptnet5/wiki/API}) for an easier extraction of per-language data. Code available on \url{https://github.com/d-gurgurov/Conceptnet-Embeddings}}. The number of triples extracted for the chosen languages are given in Table \ref{tab:language-stats}. These triples were converted into natural language using a predefined mapping from ConceptNet relationships to natural language predicates. This mapping allows for a straightforward method for injecting the graph knowledge through MLM-like objectives. The relationship mapping includes all possible connections from the ontology for the selected languages and is as specified in Table \ref{tab:relationship_mapping}.
An example of constructing a natural language sentence from an extracted triple is as follows: the triple \textit{(kiel, RelatedTo, eat)} is converted into the sentence \textit{"kiel is related to eat"}. In this context, "kiel" is the Maltese word for "eat." The natural language predicates are always kept in English, resulting in the generated text for a triple being multilingual.

\subsubsection{ConceptNet-based Language Adapters}
The ConceptNet language adapters are sequential bottleneck adapters \cite{pfeiffer2020adapterfusion}, similar to the ones used in MAD-X, with modifications to exclude invertible adapter layers for simplicity. Further, different objective functions were used for various downstream tasks at hand. For Sentiment Analysis (SA), we used the standard MLM objective, whereas for Named Entity Recognition (NER) another self-designed objective function, targeted Masked Language Modeling (TLM), was used for training the language adapters on the graph knowledge. The latter objective implies predicting the masked tokens not included in a natural language predicates specified in Table \ref{tab:relationship_mapping}. Following the earlier provided example with the sentence \textit{"kiel is related to eat"}, only either the word \textit{"kiel"} or \textit{"eat"} would be masked. 

\subsubsection{Wikipedia-based Language Adapters}
In contrast, the language adapters trained on Wikipedia data utilize the Wikimedia dataset\footnote{\url{https://huggingface.co/datasets/wikimedia/wikipedia}} for selected LRLs. This dataset provides a diverse and extensive collection of textual information scraped from Wikipedia for each language of interest. The number of articles available for each language is as in Table \ref{tab:language-stats}. The adapter architecture for Wikipedia language adapters is the same as the ConceptNet language adapters and uses the standard MLM objective function\footnote{All extracted ConceptNet data used for the language adapters, along with the language adapters themselves, can be found on HuggingFace (\url{https://huggingface.co/DGurgurov}).}.


\subsection{Fusion of Language Adapters}
In our search of enhancing multilingual LLMs for LRLs, we extend our investigation to the fusion of knowledge sources through Adapter Fusion \cite{pfeiffer2020adapterfusion}. The Adapter Fusion mechanism facilitates the integration of knowledge extracted from ConceptNet and Wikipedia-based language adapters, providing a non-destructive method to combine multiple pre-trained adapters for new downstream tasks.

An adapter fusion block introduces a set of parameters that dynamically combines adapters and the shared pre-trained model at each layer of the transformer. The fusion layer incorporates Key, Value, and Query matrices at each layer to learn contextual activation of each adapter. This dynamic combination is achieved through a contextual activation mechanism similar to attention mechanisms \cite{attention}.

We activate the fusion layer with two language adapters for each language - the Wikipedia adapter and the ConceptNet adapter. This fusion layer is introduced to allow the model to learn the optimal way to dynamically compose the knowledge from different sources. The learnable weights (Query, Key, and Value) within adapter fusion should enable the model to identify and activate the most relevant information from each adapter based on the context of the task. 

\subsection{Task Adapters}
To fine-tune multilingual LLMs for specific downstream tasks such as sentiment analysis (SA) and named entity recognition (NER), we employ task adapters stacked on top of language adapters.

The architecture of the task adapters follows the established design, featuring a stack of task-specific adapters on the top layers of a multilingual LLM and language adapter. For our study, we specifically focus on SA and NER, aiming to evaluate the impact of external knowledge injection on sentiment classification and entity recognition in LRLs, as these tasks are the most accessible in terms of labeled data availability for the selected languages.

To maintain the knowledge of the pre-trained model and language adapters during task adaptation, we adopt a weight freezing strategy, as in MAD-X. This involves preventing further fine-tuning of the weights in the pre-trained model and language adapters when training the task adapters. By doing so, we ensure that the foundational knowledge captured by the language adapters, whether sourced from Wikipedia, ConceptNet, or their fusion, remains unchanged.

The task adapters are stacked on top of the language adapters, like in the original MAD-X architecture. This stacking configuration facilitates the flow of information from the base model and the external knowledge sources through the language adapters to the task-specific adapters.

\begin{table}[t]
    \centering
    \small
    \begin{tabular}{l c>{\centering\arraybackslash}p{1.2cm}>{\centering\arraybackslash}p{1.3cm}>{\centering\arraybackslash}p{1.1cm}}
        \toprule
        \textbf{Language} & \textbf{ISO} & \textbf{CN} & \textbf{Wiki} & \textbf{mBERT?} \\
        \midrule
        Bulgarian & bg & 58060 & 297516 & \checkmark \\
        Indonesian & ms & 44190 & 689034 & \checkmark \\
        Nepali & ne & 7497 & 33040 & \checkmark \\
        Javanese & jv & 5082 & 73311 & \checkmark \\
        Maltese & mt & 8578 & 6310 & \\
        Uyghur & ug & 3225 & 5979 & \\
        Tibetan & bo & 9532 & 7090 & \\
        Sinhala & si & 3350 & 20454 & \\
        \bottomrule
    \end{tabular}
    \caption{ Number of ConceptNet triples and Wikipedia articles per language. The last column indicates if the respective language was included in the mBERT pre-training data.}
    \label{tab:language-stats}
\end{table}

\section{Experiments}
In this section, we detail the experiments and data used for conducting the study.

\begin{table*}[htb]
    \centering
    \small
    \begin{tabular}{c c c c c c c c c}
        \toprule
        \textbf{Model/Language} & \textbf{bg} & \textbf{ms} & \textbf{ne} & \textbf{jv} & \textbf{mt} & \textbf{ug} & \textbf{bo} & \textbf{si}  \\
        \midrule
        \multicolumn{9}{c}{\textbf{Sentiment Analysis (SA)}} \\
        \midrule
        mBERT & 0.860 & 0.888 & 0.565 & 0.728 & 0.557 & 0.696 & 0.687 & 0.646 \\
        mBERT+TA & 0.885 & 0.917 & 0.565 & \textbf{0.761} & 0.598 & 0.734 & 0.801 & 0.661 \\
        \\
        mBERT+Wiki+TA & \textbf{0.893} & \textbf{0.919} & \textbf{0.584} & 0.746 & \textbf{0.702} & 0.706 & \textbf{0.816} & \textbf{0.663} \\
        mBERT+CN+TA & \textbf{0.893} & 0.915 & \textbf{0.636} & 0.751 & \textbf{0.658} & 0.699 & \textbf{0.803} & 0.653 \\
        mBERT+F(CN\&Wiki)+TA & 0.882 & 0.906 & \textbf{0.627} & 0.750 & \textbf{0.662} & \textbf{0.784} & \textbf{0.804} & \textbf{0.689} \\
        \midrule
        \multicolumn{9}{c}{\textbf{Named Entity Recognition (NER)}} \\
        \midrule
        mBERT & \textbf{0.919} & \textbf{0.934} & 0.694 & \textbf{0.575} & 0.595 & 0.402 & 0.520 & 0.197 \\
        mBERT+TA & 0.917 & \textbf{0.934} & 0.644 & 0.564 & 0.601 & 0.383 & 0.575 & 0.172 \\
        \\
        mBERT+Wiki+TA & 0.915 & 0.932 & 0.610 & 0.543 & \textbf{0.603} & \textbf{0.411} & \textbf{0.576} & 0.172 \\
        mBERT+CN+TA & 0.915 & 0.928 & 0.649 & 0.571 & 0.576 & \textbf{0.403} & 0.544 & \textbf{0.244} \\
        mBERT+F(CN\&Wiki)+TA & 0.888 & 0.889 & \textbf{0.713} & 0.503 & 0.563 & 0.401 & 0.540 & 0.165 \\
        \bottomrule
    \end{tabular}
    \caption{Experimental Results. All numbers are averaged over 3 independent runs. The scores in \textbf{bold} are the ones that outperform both baseline models.}
    \label{tab:adapter-models}
\end{table*}

\subsection{Languages}
The focus languages, classified as low-resource according to \citeauthor{joshi2020state} (\citeyear{joshi2020state}), are Maltese, Bulgarian, Indonesian, Nepali, Javanese, Uyghur, Tibetan, and Sinhala, as presented in Table \ref{tab:language-stats}.
These languages serve as a subset of underrepresented languages for injecting external knowledge through ConceptNet and Wikipedia-based language adapters. The choice is bounded to the ConceptNet and downstream tasks data available for the languages. While Bulgarian, Indonesian, Nepali and Javanese data were used for pre-training mBERT \cite{devlin2018bert}, which is the multilingual LLM we will use for our experiments, Maltese, Tibetan, Uyghur and Sinhala were not included in the pre-training dataset.

\subsection{Tasks}
Two tasks considered for empirical evaluation are Sentiment Analysis (SA) and Named Entity Recognition (NER).
Datasets for SA for all the languages are acquired from different sources \cite{martinez2021evaluating, purwarianti2019improving, cortis-davis-2019-social, dingli2016sentiment, 9381292, wongso2021causal, li2022senti, 10348366, ranathunga2021sentiment} and available in our HuggingFace repositories\footnote{\url{https://huggingface.co/DGurgurov}}. For NER, the datasets are obtained from the WikiANN project \cite{pan-etal-2017-cross}. All datasets for both SA and NER contain various amounts of data, depending on language and task, and described in more detail in Appendix \ref{sec:appendixA}. We use the vanilla F1 score \cite{sokolova2006beyond} for SA performance monitoring and the "seqeval" F1 score \cite{seqeval} for NER.

While these datasets do not allow for a full assessment of the impact of injected graph knowledge on LRLs due to a lack of labeled data for other tasks, they serve as a good starting point for measuring the effects of multilingual graph knowledge integration on LRLs.

\subsection{Baselines}
In establishing baseline models for our study, we fine-tune the widely used mBERT (bert-base-multilingual-cased) \cite{devlin2018bert} from the transformer library \cite{wolf2019huggingface}. The focus is on two baseline scenarios for each task—SA and NER.

The first baseline involves fine-tuning mBERT directly on the respective datasets for SA and NER. We employ common hyperparameters for training, including a learning rate of $\{1e-4, 2e-4\}$, a batch size of $64$, $\{50, 100\}$ epochs with best-model-saving, and a dropout rate of $\{0.5, 0.2\}$, respectively for each task. The choice of hyperparameters is aligned with standard practices in transformer-based model training and our own experiments on the given datasets.

For the second baseline (mBERT+TA), we introduce a single task adapter on top of mBERT and fine-tune it on the SA and NER datasets, while keeping the parameters of mBERT frozen, using the same hyperparameters as used for the first baseline configuration. This single adapter architecture allows us to explore the effectiveness of a more compact adaptation strategy compared to the traditional fine-tuning approach.

For all models, the best checkpoint for evaluation is selected based on validation loss performance.

\subsection{Language Adapter Training}
In this section, we provide technical details of the training process for language adapters, focusing on both ConceptNet (\textit{CN}) and Wikipedia (\textit{Wiki}) variants. We use sequential bottleneck architecture without invertible layers for training language adapters.

We maintain identical hyperparameters during the training of both Wikipedia and ConceptNet language adapters: reduction factor $16$, learning rate $5e-5$, train and eval batch size $16$, training steps $50,000$ for CN and $100,000$ for Wiki. These hyperparameters ensure a stable and uniform training environment for both variants. The training process is monitored through loss and accuracy metrics on the validation sets.

\subsection{Task Adapters Training}
In this setting, we stack task-specific adapters on top of language adapters and train the task adapters while keeping the language adapters frozen. After empirical investigations, we employ similar hyperparameters to those used for training the baselines: a learning rate of $\{1e-5, 1e-4\}$, a batch size of $64$, $\{50, 100\}$ epochs with best-model saving, and a dropout of $\{0.5, 0.2\}$ for SA and NER, respectively. The experiment involves stacking the task adapters on top of either Wikipedia-based (\textit{Wiki+TA}) or ConceptNet-based (\textit{CN+TA}) language adapters, as well as on top of the fusion of both (\textit{F(CN\&Wiki)+TA}) \cite{pfeiffer2020adapterfusion}.

\begin{table*}[t]
    \centering
    \small
    \begin{tabularx}{\textwidth}{lcXXXXXXXX}
    \toprule
        \multicolumn{10}{c}{\textbf{Sentiment Analysis (SA)}}\\ \midrule
         \textbf{Configuration} & \textbf{Objective} & \textbf{bg} & \textbf{ms} & \textbf{ne} & \textbf{jv} & \textbf{mt} & \textbf{ug} & \textbf{bo} & \textbf{si} \\ \midrule
        \multirow{ 3}{*}{\textit{CN+TA}}  & MLM  & 0.893 & 0.915 & \textbf{0.636} & 0.751 & 0.658 & \textbf{0.699} & 0.803 & \textbf{0.653} \\
        & FLM & \textbf{0.898} & 0.916 & 0.575 & \textbf{0.756} & 0.639 & 0.685 & \textbf{0.811} & 0.650 \\
        & TLM & 0.893 & \textbf{0.918} & 0.625 & 0.749 & \textbf{0.661} & 0.638 & 0.797 & \textbf{0.653} \\
         \\
        \multirow{ 3}{*}{\textit{F(CN\&Wiki)+TA}}  & MLM  & 0.882 & 0.906 & 0.627 & \textbf{0.750} & 0.662 & \textbf{0.784} & 0.804 & \textbf{0.689} \\
        & FLM & 0.877 & 0.906 & \textbf{0.669} & 0.745 & 0.640 & 0.677 & 0.806 & 0.661 \\
        & TLM & \textbf{0.884} & \textbf{0.912} & 0.598 & 0.742 & \textbf{0.667} & 0.712 & \textbf{0.816} & 0.659 \\
         \midrule
         \multicolumn{10}{c}{\textbf{Named Entity Recognition (NER)}} \\
         \midrule
         \textbf{Configuration} & \textbf{Objective} & \textbf{bg} & \textbf{ms} & \textbf{ne} & \textbf{jv} & \textbf{mt} & \textbf{ug} & \textbf{bo} & \textbf{si}  \\
         \midrule
        \multirow{ 3}{*}{\textit{CN+TA}}  & MLM  & 0.915 & \textbf{0.932} & \textbf{0.657} & \textbf{0.603} & 0.471 & 0.341 & 0.559 & 0.196 \\
        & FLM & \textbf{0.918} & 0.930 & 0.626 & 0.578 & 0.476 & 0.398 & \textbf{0.572} & 0.242 \\
        & TLM & 0.915 & 0.928 & 0.649 & 0.571 & \textbf{0.576} & \textbf{0.403} & 0.544 & \textbf{0.244} \\
         \\
        \multirow{ 3}{*}{\textit{F(CN\&Wiki)+TA}}  & MLM  & \textbf{0.889} & \textbf{0.901} & 0.670 & \textbf{0.504} & 0.540 & 0.373 & 0.514 & \textbf{0.261} \\
        & FLM & 0.887 & 0.900 & 0.688 & 0.496 & \textbf{0.580} & 0.387 & 0.509 & 0.250 \\
        & TLM & 0.888 & 0.889 & \textbf{0.713} & 0.503 & 0.563 & \textbf{0.401} & \textbf{0.540} & 0.165 \\
    \bottomrule
    \end{tabularx}
    \caption{Comparison of various objective functions used for training ConceptNet based Language Adapters-Token Masked Language Modeling \textbf{(MLM)}, Full-Word Masked Language Modeling \textbf{(FLM)}, and Targeted Masked Language Modeling \textbf{(TLM)}. Maximum score per configuration in \textbf{bold}. SA results are based on MLM, and NER results are based on TLM.}
    \label{tab:adapters-ablation}
\end{table*}

\subsection{Objective Functions} \label{ablation}
In this section, we compare different objective functions used for training language adapters on graph knowledge and examine their influence on the performance on the downstream tasks at hand. We experiment with three objectives for language modeling - standard token Masked Language Modeling (MLM) \cite{devlin2018bert}, full-word Masked Language Modeling (FLM) \cite{cui2021pre}, and targeted Masked Language Modeling (TLM). MLM and FLM were implemented as provided by the Transformers library, and TLM was self-designed. MLM masks individual tokens with a 15\% probability, FLM performs the same but masks full words, and TLM masks targeted words which are not part of the natural language predicates list extracted from ConceptNet with a 50\% probability. The implied goal of TLM is to create the connections between the words of a LRL to the words of other languages, which might come in bigger quantities, within the parameters of a model. The downstream results for all the objectives are as in Table \ref{tab:adapters-ablation}. Upon the inspection of the results, different objective functions were chosen for SA and NER, according to the outcomes of the experiments. MLM was utilized as an objective for the language adapters used for SA, and TLM was chosen as an objective for the language adapters used for NER.

\section{Results and Discussion}
The experimental results, summarized in Table \ref{tab:adapter-models}, demonstrate the results of different model configurations in improving SA and NER tasks across LRLs. This section discusses the performance of each model configuration and provide insights into the impact of external knowledge through language and task adapters on enhancing multilingual LLMs for LRLs. All scores are an average over three independent runs. 

\subsection{Sentiment Analysis}

In SA, the performance of various model configurations on different languages reveals interesting insights. First, considering the baseline performance of fully fine-tuned mBERT across all languages, we observe moderate to high F1-scores. However, when single task-specific adapters are added to mBERT, we notice consistent improvements across all languages, indicating the effectiveness of using parameter-efficient fine-tuning techniques for adapting the model for specific languages, especially in low-resource scenarios. This confirms the findings by \citeauthor{li-liang-2021-prefix} (\citeyear{li-liang-2021-prefix}), \citeauthor{he-etal-2021-effectiveness} (\citeyear{he-etal-2021-effectiveness}), and \citeauthor{jukic-snajder-2023-parameter} (\citeyear{jukic-snajder-2023-parameter}).

When incorporating language adapters trained on CN and Wiki, relatively good results are observed. For nearly all languages, using language adapters trained on CN and Wiki leads to performance gains compared to the baselines-mBERT and mBERT with a single task adapter. The CN language adapter boosts the performance for Bulgarian, Nepalese, Maltese, and Tibetan over both baselines. As for the Wiki language adapters, they improve the scores for Bulgarian, Indonesian, Nepalese, Maltese, Tibetan, and Sinhala when compared to both mBERT and mBERT with a single adapter. The fusion of language adapters yields improvements over the baselines for Nepalese, Maltese, Uyghur, Tibetan, and Sinhala.

\subsection{Named Entity Recognition}
In NER, the performance trends across different model configurations and languages exhibit similar patterns to SA but with some notable differences. mBERT demonstrates moderate to high F-1 scores across languages, indicating its ability to recognize named entities to some extent. However, the addition of single task-specific adapters leads to marginal improvements in only some cases, suggesting that named entity recognition might not benefit significantly from single task-specific adapter fine-tuning compared to SA. The improvements are only observed for Maltese and Tibetan.

When incorporating language adapters, particularly those trained on CN and Wiki, we observe mixed results. Utilizing CN language adapters leads to slight improvements over the baselines only in the case of Uyghur and Sinhala. Wiki language adapters, on the other hand, give improvements over both baseline models only for Maltese, Uyghur, and Tibetan. The combination of CN and Wiki adapters shows positive impact only on Nepalese.

\subsection{Effects of Data Quantity and Language Presence in LLM pre-training Data}
The data quantity of external data sources might play a crucial role in the performance of language adapters and their impact on downstream tasks. Looking at the data quantities provided in Table \ref{tab:language-stats}, languages like Maltese, Nepali, Uyghur, Tibetan, and Sinhala have notably fewer CN and Wiki resources compared to languages like Bulgarian and Indonesian. Despite this, language adapters trained on these limited resources still contribute to performance enhancements in SA and NER tasks for these languages compared to the baseline models. This indicates the effectiveness of leveraging even small amounts of external knowledge for adapting LLMs to low-resource languages.

Another interesting observation is the performance improvement in languages like Maltese, Uyghur, Tibetan, and Sinhala, which are not included in the mBERT pre-training data. This emphasizes that the method might be more useful for languages absent in the pre-training corpus as mBERT benefits from this adaptation using task-specific and language adapters, allowing them to effectively learn from external knowledge sources and adapt to new languages.

\subsection{Take-aways}
The experimental results shed light on the effectiveness of integrating graph knowledge from linguistic ontologies into multilingual LLMs via adapters for LRLs. Across both SA and NER tasks, we observe that single task-specific adapters generally lead to performance improvements, emphasizing the benefits of parameter-efficient fine-tuning for specific tasks \cite{li-liang-2021-prefix, he-etal-2021-effectiveness, jukic-snajder-2023-parameter}.

In turn, the impact of language adapters trained on external knowledge sources such as CN and Wiki varies across languages and tasks. CN-based adapters generally show promise in enhancing SA but not NER. Wiki language adapters are also more beneficial for SA than NER.

The combination of both ConceptNet and Wikipedia adapters through the Adapter Fusion demonstrates competitive performance, in some cases outperforming individual adapters alone, suggesting that leveraging diverse knowledge sources can effectively enhance the capabilities of multilingual LLMs for low resource scenarios.

Our findings underscore the partial effectiveness of our method in leveraging external graph knowledge to enhance SA and NER tasks for individual LRLs. This highlights the need for further research to develop more effective strategies for adapting multilingual LLMs to low-resource contexts using various types of knowledge. Further, the results emphasize that each LRL needs an individual approach when building the dedicated NLP tools, where some languages might benefit from a certain method and the others might not need it.  

\section{Conclusion}
In this study, we investigated the integration of structured graph knowledge into multilingual LLMs for LRLs using language adapters and task-specific adapters. We explored the use of ConceptNet and Wikipedia data for training language adapters, and we examined Adapter Fusion as a method to combine knowledge sources. Additionally, we implemented task adapters for fine-tuning LLMs for specific downstream tasks such as Sentiment Analysis (SA) and Named Entity Recognition (NER).

Our experiments revealed insights into the effectiveness of different model configurations in improving SA and NER tasks performance across LRLs. We observed a positive effect of incorporating external graph and textual knowledge through language adapters for a number of languages, including Bulgarian, Indonesian, Maltese, Nepali, Uyghur, Tibetan, and Sinhala, some of which did not possess extensive data for training both language adapters and task adapters. Fusion of knowledge sources yielded improvements in less cases, suggesting the need for further refinement in this area.

Overall, our findings underscore the importance of parameter-efficient fine-tuning methods and the potential benefits of leveraging external knowledge for enhancing multilingual LLMs in low resource contexts. However, there are limitations to our approach, including the choice of objective functions and the need for tasks better suited to leverage external knowledge.

\section*{Limitations and Future Work}
Our approach shows several limitations that should be taken into consideration in future investigations aiming to integrate graph knowledge into multilingual LLMs for enhancing LRL performance. Firstly, the choice of objective function employed for learning graph knowledge plays a critical role in effectively acquiring underlying knowledge. The objectives we explored may not be optimally suited for this purpose, highlighting the need for more tailored approaches to graph knowledge acquisition. Secondly, the tasks we selected for evaluating the effectiveness of knowledge injection may not inherently require the type of knowledge provided by graph sources. Future work should explore tasks that better leverage the acquired knowledge. Thirdly, our study was limited to a subset of LRLs, and expanding the scope to include a broader range of languages would provide a more comprehensive assessment of our approach's effectiveness. Lastly, larger models should be explored as backbones to build upon.

\section*{Acknowledgments}
We are thankful to the anonymous reviewers for their insightful comments and suggestions. This research was supported by the EU-funded LT-Bridge project, GA No. 952194; and the EU-funded project DisAI - Improving scientific excellence and creativity in combating disinformation with artificial intelligence and language technologies, GA No. 101079164. Additionally, this work received support from the German Federal Ministry of Education and Research as part of the TRAILS project (GA No. 01IW24005). 

\bibliography{custom}

\begin{thebibliography}{39}
\expandafter\ifx\csname natexlab\endcsname\relax\def\natexlab#1{#1}\fi

\bibitem[{Artetxe et~al.(2020)Artetxe, Ruder, and Yogatama}]{artetxe-etal-2020-cross}
Mikel Artetxe, Sebastian Ruder, and Dani Yogatama. 2020.
\newblock \href {https://doi.org/10.18653/v1/2020.acl-main.421} {On the cross-lingual transferability of monolingual representations}.
\newblock In \emph{Proceedings of the 58th Annual Meeting of the Association for Computational Linguistics}, pages 4623--4637, Online. Association for Computational Linguistics.

\bibitem[{Conneau et~al.(2020)Conneau, Khandelwal, Goyal, Chaudhary, Wenzek, Guzm{\'a}n, Grave, Ott, Zettlemoyer, and Stoyanov}]{conneau2019unsupervised}
Alexis Conneau, Kartikay Khandelwal, Naman Goyal, Vishrav Chaudhary, Guillaume Wenzek, Francisco Guzm{\'a}n, Edouard Grave, Myle Ott, Luke Zettlemoyer, and Veselin Stoyanov. 2020.
\newblock \href {https://doi.org/10.18653/v1/2020.acl-main.747} {Unsupervised cross-lingual representation learning at scale}.
\newblock In \emph{Proceedings of the 58th Annual Meeting of the Association for Computational Linguistics}, pages 8440--8451, Online. Association for Computational Linguistics.

\bibitem[{Cortis and Davis(2019)}]{cortis-davis-2019-social}
Keith Cortis and Brian Davis. 2019.
\newblock \href {https://doi.org/10.18653/v1/D19-5547} {A social opinion gold standard for the {M}alta government budget 2018}.
\newblock In \emph{Proceedings of the 5th Workshop on Noisy User-generated Text (W-NUT 2019)}, pages 364--369, Hong Kong, China. Association for Computational Linguistics.

\bibitem[{Cui et~al.(2021)Cui, Che, Liu, Qin, and Yang}]{cui2021pre}
Yiming Cui, Wanxiang Che, Ting Liu, Bing Qin, and Ziqing Yang. 2021.
\newblock Pre-training with whole word masking for chinese bert.
\newblock \emph{IEEE/ACM Transactions on Audio, Speech, and Language Processing}, 29:3504--3514.

\bibitem[{Devlin et~al.(2019)Devlin, Chang, Lee, and Toutanova}]{devlin2018bert}
Jacob Devlin, Ming-Wei Chang, Kenton Lee, and Kristina Toutanova. 2019.
\newblock \href {https://doi.org/10.18653/v1/N19-1423} {{BERT}: Pre-training of deep bidirectional transformers for language understanding}.
\newblock In \emph{Proceedings of the 2019 Conference of the North {A}merican Chapter of the Association for Computational Linguistics: Human Language Technologies, Volume 1 (Long and Short Papers)}, pages 4171--4186, Minneapolis, Minnesota. Association for Computational Linguistics.

\bibitem[{Dingli and Sant(2016)}]{dingli2016sentiment}
Alexiei Dingli and Nicole Sant. 2016.
\newblock Sentiment analysis on maltese using machine learning.
\newblock In \emph{Proceedings of The Tenth International Conference on Advances in Semantic Processing (SEMAPRO 2016)}, pages 21--25.

\bibitem[{Hangya et~al.(2022)Hangya, Saadi, and Fraser}]{hangya-etal-2022-improving}
Viktor Hangya, Hossain~Shaikh Saadi, and Alexander Fraser. 2022.
\newblock \href {https://doi.org/10.18653/v1/2022.emnlp-main.822} {Improving low-resource languages in pre-trained multilingual language models}.
\newblock In \emph{Proceedings of the 2022 Conference on Empirical Methods in Natural Language Processing}, pages 11993--12006, Abu Dhabi, United Arab Emirates. Association for Computational Linguistics.

\bibitem[{He et~al.(2021)He, Liu, Ye, Tan, Ding, Cheng, Low, Bing, and Si}]{he-etal-2021-effectiveness}
Ruidan He, Linlin Liu, Hai Ye, Qingyu Tan, Bosheng Ding, Liying Cheng, Jiawei Low, Lidong Bing, and Luo Si. 2021.
\newblock \href {https://doi.org/10.18653/v1/2021.acl-long.172} {On the effectiveness of adapter-based tuning for pretrained language model adaptation}.
\newblock In \emph{Proceedings of the 59th Annual Meeting of the Association for Computational Linguistics and the 11th International Joint Conference on Natural Language Processing (Volume 1: Long Papers)}, pages 2208--2222, Online. Association for Computational Linguistics.

\bibitem[{Hou et~al.(2022)Hou, Jiao, Liu, Allen, Tu, and Sachan}]{hou2022adapters}
Yifan Hou, Wenxiang Jiao, Meizhen Liu, Carl Allen, Zhaopeng Tu, and Mrinmaya Sachan. 2022.
\newblock Adapters for enhanced modeling of multilingual knowledge and text.
\newblock In \emph{Findings of the Association for Computational Linguistics: EMNLP 2022}, pages 3902--3917.

\bibitem[{Houlsby et~al.(2019)Houlsby, Giurgiu, Jastrzebski, Morrone, De~Laroussilhe, Gesmundo, Attariyan, and Gelly}]{houlsby2019parameter}
Neil Houlsby, Andrei Giurgiu, Stanislaw Jastrzebski, Bruna Morrone, Quentin De~Laroussilhe, Andrea Gesmundo, Mona Attariyan, and Sylvain Gelly. 2019.
\newblock Parameter-efficient transfer learning for nlp.
\newblock In \emph{International Conference on Machine Learning}, pages 2790--2799. PMLR.

\bibitem[{Joshi et~al.(2020)Joshi, Santy, Budhiraja, Bali, and Choudhury}]{joshi2020state}
Pratik Joshi, Sebastin Santy, Amar Budhiraja, Kalika Bali, and Monojit Choudhury. 2020.
\newblock \href {https://doi.org/10.18653/v1/2020.acl-main.560} {The state and fate of linguistic diversity and inclusion in the {NLP} world}.
\newblock In \emph{Proceedings of the 58th Annual Meeting of the Association for Computational Linguistics}, pages 6282--6293, Online. Association for Computational Linguistics.

\bibitem[{Juki{\'c} and Snajder(2023)}]{jukic-snajder-2023-parameter}
Josip Juki{\'c} and Jan Snajder. 2023.
\newblock \href {https://doi.org/10.18653/v1/2023.emnlp-main.307} {Parameter-efficient language model tuning with active learning in low-resource settings}.
\newblock In \emph{Proceedings of the 2023 Conference on Empirical Methods in Natural Language Processing}, pages 5061--5074, Singapore. Association for Computational Linguistics.

\bibitem[{Lauscher et~al.(2020)Lauscher, Majewska, Ribeiro, Gurevych, Rozanov, and Glava{\v{s}}}]{lauscher2020common}
Anne Lauscher, Olga Majewska, Leonardo F.~R. Ribeiro, Iryna Gurevych, Nikolai Rozanov, and Goran Glava{\v{s}}. 2020.
\newblock \href {https://doi.org/10.18653/v1/2020.deelio-1.5} {Common sense or world knowledge? investigating adapter-based knowledge injection into pretrained transformers}.
\newblock In \emph{Proceedings of Deep Learning Inside Out (DeeLIO): The First Workshop on Knowledge Extraction and Integration for Deep Learning Architectures}, pages 43--49, Online. Association for Computational Linguistics.

\bibitem[{Li et~al.(2022)Li, Zhao, Yang, Jiang, Li, and Ma}]{li2022senti}
Siyu Li, Kui Zhao, Jin Yang, Xinyun Jiang, Zhengji Li, and Zicheng Ma. 2022.
\newblock Senti-exlm: Uyghur enhanced sentiment analysis model based on xlm.
\newblock \emph{Electronics Letters}, 58(13):517--519.

\bibitem[{Li and Liang(2021)}]{li-liang-2021-prefix}
Xiang~Lisa Li and Percy Liang. 2021.
\newblock \href {https://doi.org/10.18653/v1/2021.acl-long.353} {Prefix-tuning: Optimizing continuous prompts for generation}.
\newblock In \emph{Proceedings of the 59th Annual Meeting of the Association for Computational Linguistics and the 11th International Joint Conference on Natural Language Processing (Volume 1: Long Papers)}, pages 4582--4597, Online. Association for Computational Linguistics.

\bibitem[{Liu et~al.(2019)Liu, Ott, Goyal, Du, Joshi, Chen, Levy, Lewis, Zettlemoyer, and Stoyanov}]{liu2019roberta}
Yinhan Liu, Myle Ott, Naman Goyal, Jingfei Du, Mandar Joshi, Danqi Chen, Omer Levy, Mike Lewis, Luke Zettlemoyer, and Veselin Stoyanov. 2019.
\newblock \href {http://arxiv.org/abs/1907.11692} {Roberta: {A} robustly optimized {BERT} pretraining approach}.
\newblock \emph{CoRR}, abs/1907.11692.

\bibitem[{Mart{\'\i}nez-Garc{\'\i}a et~al.(2021)Mart{\'\i}nez-Garc{\'\i}a, Badia, and Barnes}]{martinez2021evaluating}
Antonio Mart{\'\i}nez-Garc{\'\i}a, Toni Badia, and Jeremy Barnes. 2021.
\newblock Evaluating morphological typology in zero-shot cross-lingual transfer.
\newblock In \emph{Proceedings of the 59th Annual Meeting of the Association for Computational Linguistics and the 11th International Joint Conference on Natural Language Processing (Volume 1: Long Papers)}, pages 3136--3153.

\bibitem[{Miller(1995)}]{miller1995wordnet}
George~A Miller. 1995.
\newblock Wordnet: a lexical database for english.
\newblock \emph{Communications of the ACM}, 38(11):39--41.

\bibitem[{Muller et~al.(2021)Muller, Anastasopoulos, Sagot, and Seddah}]{muller-etal-2021-unseen}
Benjamin Muller, Antonios Anastasopoulos, Beno{\^\i}t Sagot, and Djam{\'e} Seddah. 2021.
\newblock \href {https://doi.org/10.18653/v1/2021.naacl-main.38} {When being unseen from m{BERT} is just the beginning: Handling new languages with multilingual language models}.
\newblock In \emph{Proceedings of the 2021 Conference of the North American Chapter of the Association for Computational Linguistics: Human Language Technologies}, pages 448--462, Online. Association for Computational Linguistics.

\bibitem[{Nakayama(2018)}]{seqeval}
Hiroki Nakayama. 2018.
\newblock \href {https://github.com/chakki-works/seqeval} {{seqeval}: A python framework for sequence labeling evaluation}.
\newblock Software available from https://github.com/chakki-works/seqeval.

\bibitem[{Pan et~al.(2017)Pan, Zhang, May, Nothman, Knight, and Ji}]{pan-etal-2017-cross}
Xiaoman Pan, Boliang Zhang, Jonathan May, Joel Nothman, Kevin Knight, and Heng Ji. 2017.
\newblock \href {https://doi.org/10.18653/v1/P17-1178} {Cross-lingual name tagging and linking for 282 languages}.
\newblock In \emph{Proceedings of the 55th Annual Meeting of the Association for Computational Linguistics (Volume 1: Long Papers)}, pages 1946--1958, Vancouver, Canada. Association for Computational Linguistics.

\bibitem[{Pfeiffer et~al.(2022)Pfeiffer, Goyal, Lin, Li, Cross, Riedel, and Artetxe}]{pfeiffer-etal-2022-lifting}
Jonas Pfeiffer, Naman Goyal, Xi~Lin, Xian Li, James Cross, Sebastian Riedel, and Mikel Artetxe. 2022.
\newblock \href {https://doi.org/10.18653/v1/2022.naacl-main.255} {Lifting the curse of multilinguality by pre-training modular transformers}.
\newblock In \emph{Proceedings of the 2022 Conference of the North American Chapter of the Association for Computational Linguistics: Human Language Technologies}, pages 3479--3495, Seattle, United States. Association for Computational Linguistics.

\bibitem[{Pfeiffer et~al.(2021)Pfeiffer, Kamath, R{\"u}ckl{\'e}, Cho, and Gurevych}]{pfeiffer2020adapterfusion}
Jonas Pfeiffer, Aishwarya Kamath, Andreas R{\"u}ckl{\'e}, Kyunghyun Cho, and Iryna Gurevych. 2021.
\newblock \href {https://doi.org/10.18653/v1/2021.eacl-main.39} {{A}dapter{F}usion: Non-destructive task composition for transfer learning}.
\newblock In \emph{Proceedings of the 16th Conference of the European Chapter of the Association for Computational Linguistics: Main Volume}, pages 487--503, Online. Association for Computational Linguistics.

\bibitem[{Pfeiffer et~al.(2020)Pfeiffer, Vuli{\'c}, Gurevych, and Ruder}]{pfeiffer2020mad}
Jonas Pfeiffer, Ivan Vuli{\'c}, Iryna Gurevych, and Sebastian Ruder. 2020.
\newblock \href {https://doi.org/10.18653/v1/2020.emnlp-main.617} {{MAD-X}: {A}n {A}dapter-{B}ased {F}ramework for {M}ulti-{T}ask {C}ross-{L}ingual {T}ransfer}.
\newblock In \emph{Proceedings of the 2020 Conference on Empirical Methods in Natural Language Processing (EMNLP)}, pages 7654--7673, Online. Association for Computational Linguistics.

\bibitem[{Purwarianti and Crisdayanti(2019)}]{purwarianti2019improving}
Ayu Purwarianti and Ida Ayu Putu~Ari Crisdayanti. 2019.
\newblock Improving bi-lstm performance for indonesian sentiment analysis using paragraph vector.
\newblock In \emph{2019 International Conference of Advanced Informatics: Concepts, Theory and Applications (ICAICTA)}, pages 1--5. IEEE.

\bibitem[{Ranathunga and Liyanage(2021)}]{ranathunga2021sentiment}
Surangika Ranathunga and Isuru~Udara Liyanage. 2021.
\newblock Sentiment analysis of sinhala news comments.
\newblock \emph{Transactions on Asian and Low-Resource Language Information Processing}, 20(4):1--23.

\bibitem[{Singh et~al.(2020)Singh, Timilsina, Bal, and Joshi}]{9381292}
Oyesh~Mann Singh, Sandesh Timilsina, Bal~Krishna Bal, and Anupam Joshi. 2020.
\newblock \href {https://doi.org/10.1109/ASONAM49781.2020.9381292} {Aspect based abusive sentiment detection in nepali social media texts}.
\newblock In \emph{2020 IEEE/ACM International Conference on Advances in Social Networks Analysis and Mining (ASONAM)}, pages 301--308.

\bibitem[{Singh et~al.(2002)Singh, Lin, Mueller, Lim, Perkins, and Li~Zhu}]{singh2002open}
Push Singh, Thomas Lin, Erik~T Mueller, Grace Lim, Travell Perkins, and Wan Li~Zhu. 2002.
\newblock Open mind common sense: Knowledge acquisition from the general public.
\newblock In \emph{On the Move to Meaningful Internet Systems 2002: CoopIS, DOA, and ODBASE: Confederated International Conferences CoopIS, DOA, and ODBASE 2002 Proceedings}, pages 1223--1237. Springer.

\bibitem[{Sokolova et~al.(2006)Sokolova, Japkowicz, and Szpakowicz}]{sokolova2006beyond}
Marina Sokolova, Nathalie Japkowicz, and Stan Szpakowicz. 2006.
\newblock Beyond accuracy, f-score and roc: a family of discriminant measures for performance evaluation.
\newblock In \emph{Australasian joint conference on artificial intelligence}, pages 1015--1021. Springer.

\bibitem[{Speer et~al.(2017)Speer, Chin, and Havasi}]{speer2017conceptnet}
Robyn Speer, Joshua Chin, and Catherine Havasi. 2017.
\newblock Conceptnet 5.5: An open multilingual graph of general knowledge.
\newblock In \emph{Proceedings of the AAAI conference on artificial intelligence}, volume~31.

\bibitem[{Vaswani et~al.(2017)Vaswani, Shazeer, Parmar, Uszkoreit, Jones, Gomez, Kaiser, and Polosukhin}]{attention}
Ashish Vaswani, Noam Shazeer, Niki Parmar, Jakob Uszkoreit, Llion Jones, Aidan~N. Gomez, Lukasz Kaiser, and Illia Polosukhin. 2017.
\newblock \href {http://arxiv.org/abs/1706.03762} {Attention is all you need}.
\newblock \emph{CoRR}, abs/1706.03762.

\bibitem[{Vernikos and Popescu-Belis(2021)}]{vernikos-popescu-belis-2021-subword-mapping}
Giorgos Vernikos and Andrei Popescu-Belis. 2021.
\newblock \href {https://doi.org/10.18653/v1/2021.findings-emnlp.224} {Subword mapping and anchoring across languages}.
\newblock In \emph{Findings of the Association for Computational Linguistics: EMNLP 2021}, pages 2633--2647, Punta Cana, Dominican Republic. Association for Computational Linguistics.

\bibitem[{Vrande{\v{c}}i{\'c} and Kr{\"o}tzsch(2014)}]{vrandevcic2014wikidata}
Denny Vrande{\v{c}}i{\'c} and Markus Kr{\"o}tzsch. 2014.
\newblock Wikidata: a free collaborative knowledgebase.
\newblock \emph{Communications of the ACM}, 57(10):78--85.

\bibitem[{Wang et~al.(2021)Wang, Tang, Duan, Wei, Huang, Ji, Cao, Jiang, and Zhou}]{wang2020k}
Ruize Wang, Duyu Tang, Nan Duan, Zhongyu Wei, Xuanjing Huang, Jianshu Ji, Guihong Cao, Daxin Jiang, and Ming Zhou. 2021.
\newblock \href {https://doi.org/10.18653/v1/2021.findings-acl.121} {{K-Adapter}: {I}nfusing {K}nowledge into {P}re-{T}rained {M}odels with {A}dapters}.
\newblock In \emph{Findings of the Association for Computational Linguistics: ACL-IJCNLP 2021}, pages 1405--1418, Online. Association for Computational Linguistics.

\bibitem[{Wolf et~al.(2020)Wolf, Debut, Sanh, Chaumond, Delangue, Moi, Cistac, Rault, Louf, Funtowicz, Davison, Shleifer, von Platen, Ma, Jernite, Plu, Xu, Le~Scao, Gugger, Drame, Lhoest, and Rush}]{wolf2019huggingface}
Thomas Wolf, Lysandre Debut, Victor Sanh, Julien Chaumond, Clement Delangue, Anthony Moi, Pierric Cistac, Tim Rault, Remi Louf, Morgan Funtowicz, Joe Davison, Sam Shleifer, Patrick von Platen, Clara Ma, Yacine Jernite, Julien Plu, Canwen Xu, Teven Le~Scao, Sylvain Gugger, Mariama Drame, Quentin Lhoest, and Alexander Rush. 2020.
\newblock \href {https://doi.org/10.18653/v1/2020.emnlp-demos.6} {Transformers: State-of-the-art natural language processing}.
\newblock In \emph{Proceedings of the 2020 Conference on Empirical Methods in Natural Language Processing: System Demonstrations}, pages 38--45, Online. Association for Computational Linguistics.

\bibitem[{Wongso et~al.(2021)Wongso, Setiawan, and Suhartono}]{wongso2021causal}
Wilson Wongso, David~Samuel Setiawan, and Derwin Suhartono. 2021.
\newblock Causal and masked language modeling of javanese language using transformer-based architectures.
\newblock In \emph{2021 International Conference on Advanced Computer Science and Information Systems (ICACSIS)}, pages 1--7. IEEE.

\bibitem[{Wu and Dredze(2020)}]{wu-dredze-2020-languages}
Shijie Wu and Mark Dredze. 2020.
\newblock \href {https://doi.org/10.18653/v1/2020.repl4nlp-1.16} {Are all languages created equal in multilingual {BERT}?}
\newblock In \emph{Proceedings of the 5th Workshop on Representation Learning for NLP}, pages 120--130, Online. Association for Computational Linguistics.

\bibitem[{Xue et~al.(2021)Xue, Constant, Roberts, Kale, Al-Rfou, Siddhant, Barua, and Raffel}]{xue2020mt5}
Linting Xue, Noah Constant, Adam Roberts, Mihir Kale, Rami Al-Rfou, Aditya Siddhant, Aditya Barua, and Colin Raffel. 2021.
\newblock \href {https://doi.org/10.18653/v1/2021.naacl-main.41} {m{T}5: A massively multilingual pre-trained text-to-text transformer}.
\newblock In \emph{Proceedings of the 2021 Conference of the North American Chapter of the Association for Computational Linguistics: Human Language Technologies}, pages 483--498, Online. Association for Computational Linguistics.

\bibitem[{Zhu et~al.(2023)Zhu, Luosai, Zhou, Qun, and Nyima}]{10348366}
Yulei Zhu, Baima Luosai, Liyuan Zhou, Nuo Qun, and Tashi Nyima. 2023.
\newblock \href {https://doi.org/10.1109/PRML59573.2023.10348366} {Research on sentiment analysis of tibetan short text based on dual-channel hybrid neural network}.
\newblock In \emph{2023 IEEE 4th International Conference on Pattern Recognition and Machine Learning (PRML)}, pages 377--384.

\end{thebibliography}

\appendix
\section*{Appendix}
\section{SA and NER Data Details}
\label{sec:appendixA}
Table \ref{tab:sa-sources} and \ref{tab:ner-sources} provide a more detailed description of the datasets used for training task adapters. 

\begin{table}[h]
    \centering
    \small
    \begin{tabular}{l c c | r r | r r r}
        \toprule
        \textbf{Language} & \textbf{ISO code} & \textbf{Source} & \textbf{\#pos} & \textbf{\#neg} & \textbf{\#train} & \textbf{\#val} & \textbf{\#test} \\
        \midrule
        Bulgarian & bg & \citeauthor {martinez2021evaluating}, \citeyear{martinez2021evaluating} & 6652 & 1271 & 5412 & 838 & 1673 \\
        Indonesian & ms & \citeauthor{purwarianti2019improving}, \citeyear{purwarianti2019improving} & 7319 & 4005 & 7926 & 1132 & 2266 \\
        Maltese & mt & \citeauthor{cortis-davis-2019-social}, \citeyear{cortis-davis-2019-social}; \citeauthor{dingli2016sentiment}, \citeyear{dingli2016sentiment} & 271 & 580 & 595 & 85 & 171 \\
        Nepali & ne & \citeauthor{9381292}, \citeyear{9381292} & 680 & 1019 & 1189 & 255 & 255 \\
        Javanese & jv & \citeauthor{wongso2021causal}, \citeyear{wongso2021causal} & 12500 & 12500 & 17500 & 5025 & 2475 \\
        Uyghur & ug & \citeauthor{li2022senti}, \citeyear{li2022senti} & 2450 & 353 & 1962 & 311 & 530 \\
        Tibetan & bo & \citeauthor{10348366}, \citeyear{10348366} & 5006 & 5000 & 7004 & 1501 & 1501 \\
        Sinhala & si & \citeauthor{ranathunga2021sentiment}, \citeyear{ranathunga2021sentiment} & 2487 & 2516 & 3502 & 750 & 751 \\
        \bottomrule
    \end{tabular}
    \caption{Sentiment Analysis Data Details}
    \label{tab:sa-sources}
\end{table}

\begin{table}[h]
    \centering
    \small
    \begin{tabular}{l c | r r r}
        \toprule
        \textbf{Language} & \textbf{ISO code} & \textbf{\#train} & \textbf{\#val} & \textbf{\#test} \\
        \midrule
        Bulgarian & bg & 20000 & 10000 & 10000 \\
        Indonesian & ms & 20000 & 1000 & 1000 \\
        Maltese & mt & 100 & 100 & 100 \\
        Nepali & ne & 100 & 100 & 100 \\
        Javanese & jv & 100 & 100 & 100 \\
        Uyghur & ug & 100 & 100 & 100 \\
        Tibetan & bo & 100 & 100 & 100 \\
        Sinhala & si & 100 & 100 & 100 \\
        \bottomrule
    \end{tabular}
    \caption{Named Entity Recognition Data Details}
    \label{tab:ner-sources}
\end{table}

\end{document}